\documentclass{article}

\usepackage[preprint,nonatbib]{neurips_2024}

\usepackage[numbers]{natbib}

\usepackage[utf8]{inputenc} % allow utf-8 input
\usepackage[T1]{fontenc}    % use 8-bit T1 fonts
\usepackage{hyperref}       % hyperlinks
\usepackage{url}            % simple URL typesetting
\usepackage{booktabs}       % professional-quality tables
\usepackage{amsfonts}       % blackboard math symbols
\usepackage{nicefrac}       % compact symbols for 1/2, etc.
\usepackage{microtype}      % microtypography
\usepackage{xcolor}         % colors
\usepackage{amsmath}
\usepackage{cleveref}

\usepackage{amsmath}  % for align
\usepackage{physics}  % pdv
\newtheorem{theorem}{Theorem}
\newtheorem{lemma}{Lemma}
\newtheorem{proposition}{Proposition}
\usepackage{graphicx}  % includegraphics
\usepackage{multirow}
\usepackage{wrapfig}
\usepackage{subfig}
\usepackage{colortbl}  % rowcolor
\usepackage{accents}  % dbtilde

\newcommand{\ie}{\textit{i}.\textit{e}.}

\newcommand{\bI}{\mathbf{I}}
\newcommand{\bW}{\mathbf{W}}
\newcommand{\bX}{\mathbf{X}}
\newcommand{\bP}{\mathbf{P}}
\newcommand{\bXtr}{\mathbf{X}_{\textrm{train}}}
\newcommand{\bXte}{\mathbf{X}_{\textrm{test}}}
\newcommand{\bPtr}{\mathbf{P}_{\textrm{train}}}
\newcommand{\bPte}{\mathbf{P}_{\textrm{test}}}
\newcommand{\bItr}{\mathbf{I}_{\textrm{train}}}
\newcommand{\bIte}{\mathbf{I}_{\textrm{test}}}
\newcommand{\bv}{\mathbf{v}}
\newcommand{\tbv}{\tilde{\mathbf{v}}}
\newcommand{\N}{\mathcal{N}}
\newcommand{\bM}{\mathbf{M}}
\newcommand{\bbv}{\bar{\mathbf{v}}}
\newcommand{\hbv}{\hat{\mathbf{v}}}
\newcommand{\be}{\mathbf{e}}
\newcommand{\bone}{\mathbf{1}}
\newcommand{\bzero}{\mathbf{0}}

\newcommand{\dbtilde}[1]{\accentset{\approx}{#1}}
\newcommand{\dtbv}{\dbtilde{\mathbf{v}}}

\def\figref#1{Figure~\ref{#1}}
\def\tabref#1{Table~\ref{#1}}
\def\eqref#1{Eq.~\ref{#1}}

\DeclareMathOperator{\E}{E}
\DeclareMathOperator{\Var}{Var}
\DeclareMathOperator{\LN}{LN}
\DeclareMathOperator{\UP}{UP}
\DeclareMathOperator{\Crop}{Crop}

\crefname{equation}{Eq.}{Eqs.} % capitalize "E"

\makeatletter
\renewcommand{\paragraph}{%
  \@startsection{paragraph}{4}%
  {\z@}{0.00ex \@plus 1ex \@minus .2ex}{-1em}%
  {\normalfont\normalsize\bfseries}%
}
\makeatother

\makeatletter
\renewcommand\subsubsection{\@startsection{subsubsection}{3}{\z@}%
                       {-8\p@ \@plus -4\p@ \@minus -4\p@}% Formerly -18\p@ \@plus -4\p@ \@minus -4\p@
                       {-0.5em \@plus -0.22em \@minus -0.1em}%
                       {\normalfont\normalsize\bfseries\boldmath}}
\makeatother

\title{Configuring Data Augmentations to Reduce \\ Variance Shift in Positional Embedding \\ of Vision Transformers}

\author{%
  Bum Jun Kim \\
  POSTECH \\
  \texttt{kmbmjn@postech.edu} \\
  \And
  Sang Woo Kim \\
  POSTECH \\
  \texttt{swkim@postech.edu} \\
}

\begin{document}

\maketitle

\begin{abstract}
	Vision transformers (ViTs) have demonstrated remarkable performance in a variety of vision tasks. Despite their promising capabilities, training a ViT requires a large amount of diverse data. Several studies empirically found that using rich data augmentations, such as Mixup, Cutmix, and random erasing, is critical to the successful training of ViTs. Now, the use of rich data augmentations has become a standard practice in the current state. However, we report a vulnerability to this practice: Certain data augmentations such as Mixup cause a variance shift in the positional embedding of ViT, which has been a hidden factor that degrades the performance of ViT during the test phase. We claim that achieving a stable effect from positional embedding requires a specific condition on the image, which is often broken for the current data augmentation methods. We provide a detailed analysis of this problem as well as the correct configuration for these data augmentations to remove the side effects of variance shift. Experiments showed that adopting our guidelines improves the performance of ViTs compared with the current configuration of data augmentations.
\end{abstract}

\section{Introduction}
Vision transformers (ViTs) \citep{DBLP:conf/iclr/DosovitskiyB0WZ21} have demonstrated remarkable performance in a wide range of vision tasks. They have replaced the dominance of previous convolutional neural networks (CNNs) \citep{DBLP:journals/corr/SimonyanZ14a,DBLP:conf/cvpr/HeZRS16}, exhibiting improved modeling ability. Now, ViTs have become the standard backbone in image classification as well as other downstream tasks, such as semantic segmentation.

Compared with CNNs, training ViT requires a much larger or more diverse dataset. The original study of ViT \citep{DBLP:conf/iclr/DosovitskiyB0WZ21} observed that directly training ViTs on ImageNet\footnote{In this paper, ImageNet indicates ImageNet-1K \citep{DBLP:conf/cvpr/DengDSLL009} unless specified otherwise.} dataset exhibits worse results than CNNs. To overcome this problem, they pretrained ViTs on the JFT-300M dataset, which is their in-house dataset with a larger size. Subsequently, they fine-tuned the ViTs on the ImageNet dataset, which then showed improved performance compared with CNNs. Afterwards, other follow-up studies \citep{DBLP:conf/icml/TouvronCDMSJ21,DBLP:conf/eccv/TouvronCJ22} attempted to train ViTs using only ImageNet and found that successful training of ViTs requires rich data augmentations, including Mixup \citep{DBLP:conf/iclr/ZhangCDL18}, Cutmix \citep{DBLP:conf/iccv/YunHCOYC19}, and random erasing \citep{DBLP:conf/aaai/Zhong0KL020}. Since then, using a larger dataset or rich data augmentations has been discussed as a critical factor in training ViTs to achieve successful performance.

Note that, in fact, these data augmentations were developed during the CNN era. Although Mixup, Cutmix, and random erasing provide rich transformations of data with increased diversity, we should rigorously examine their validity for training ViTs. Eventually, we report a vulnerability for this issue: Certain data augmentations cause a variance shift in the positional embedding of ViT, which leads to inconsistent behavior of the positional embedding and degraded performance of ViT. We discover that this vulnerability is incurred by the distinct architecture in the early stage of ViT, which requires specific conditions for the input image.

In this regard, we inspect whether current data augmentations satisfy this condition. We provide a detailed theoretical description of this vulnerability as well as its root cause. Our analysis discovers that Cutmix is safe from this vulnerability, whereas Mixup incurs variance shifts. We also explore other data augmentations such as random erasing and random resize crop, which require specific configurations for the correct use on ViTs. We empirically examined the validity of our guidelines for configuring data augmentations and observed that following them improves the performance of ViTs.

\section{Theory and Practice: Data Augmentations for Vision Transformers}
\subsection{Problem Statement: Variance Shift in Positional Embedding}
This study targets ViTs with absolute positional embedding, which is widely used in numerous studies \citep{DBLP:conf/iccv/TouvronCSSJ21,DBLP:conf/icml/dAscoliTLMBS21,DBLP:conf/nips/JiangHYZSJWF21,DBLP:conf/nips/AliTCBDJLNSVJ21,DBLP:conf/nips/HanXWGXW21,DBLP:conf/iccv/HeoYHCCO21}. In the early stage, ViT partitions an input image $\bI$ into a sequence of patches, which is subsequently modeled by transformer blocks including self-attention operations \citep{DBLP:conf/nips/VaswaniSPUJGKP17}. The sequence of partitioned patches $\bX$---also referred to as patch embedding---is a linear projection of the input image with respect to patch, \ie, $\bX = \bW * \bI$, where * indicates a strided convolutional operation and $\bW$ indicates the convolutional kernel. Because the self-attention operation does not recognize positional differences, ViT explicitly employs positional embedding to discriminate between patches at each location. The positional embedding $\bP$ is added to the patch embedding and is set to be learnable during training. Now the transformer block starts and applies numerous operations including layer normalization (LN) \citep{DBLP:journals/corr/BaKH16}, multiheaded self-attention, and multilayer perceptron. The first operation of the transformer blocks is LN, which applies normalization using mean and standard deviation to the sum of patch and positional embeddings, \ie, $\bX+\bP$.

\begin{figure}[t!]
	\centering
	\includegraphics[width=1.0\textwidth]{"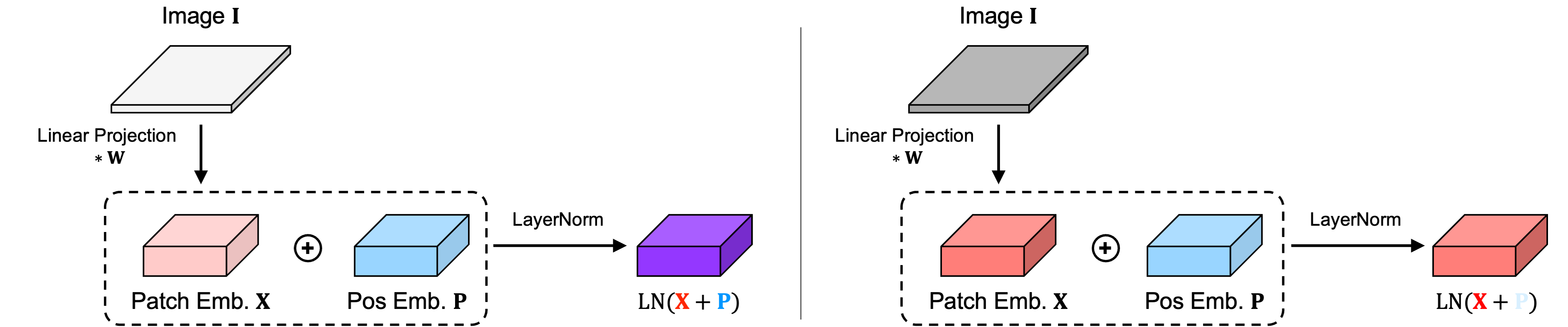"}
	\caption{Overview of the early stage of ViT. Variance is depicted by chroma. If the input image or patch embedding exhibits different variances during the training (left) and test (right) phases, positional embedding inconsistently affects the output.}
	\label{fig:emb}
\end{figure}

Owing to the first LN, the contribution of positional embedding $\bP$ depends on its relative variance with respect to the operand $\bX$ (\figref{fig:emb}). For example, if $\Var[\bX] > \Var[\bP]$, patch embedding is far more dominant than positional embedding, which decreases the contribution of positional embedding. Specifically, the gradient $\pdv{\LN(\bX+\bP)}{P}$ decays with a factor of $\sqrt{\Var[\bX]}$, which indicates that the larger variance of the patch embedding reduces the contribution of the positional embedding \citep{DBLP:journals/pr/KimCJLJK23}. Similarly, a larger variance of the positional embedding decreases the relative contribution of the patch embedding, which is equivalent to reducing the variance of the patch embedding. To ensure a consistent effect from positional embedding, we should avoid inconsistent variance for both patch and positional embeddings.

We extend this rule with respect to the training and test phases: If there is a variance shift in patch or positional embeddings during the training and test phases, ViT exhibits inconsistent behavior. To obtain consistency in the relative contribution of positional embedding, we should ensure a consistent ratio in variance $\Var[\bXtr] / \Var[\bPtr] \approx \Var[\bXte] / \Var[\bPte]$ as much as possible. Note that patch embedding is a linear projection of the input image; achieving a consistent variance on patch embedding $\Var[\bX]$ is equivalent to ensuring a consistent variance on the input image $\Var[\bI]$. Therefore, the term $\bX$ in the above ratios can be replaced with $\bI$.

However, most data augmentations are only applied during the training phase and turned off during the test phase, which can cause the vulnerability of variance shifts in the input image. In other words, data augmentations do not always guarantee a consistent variance of the input image, which breaks the aforementioned condition. Because modern data augmentations have been developed considering CNNs, which do not employ positional embedding and are free from the variance shift issue, we claim that data augmentations should be configured to be suitable for ViT. Considering this issue, our study investigates the validity of modern data augmentations as well as their correct configurations to avoid variance shifts.

These arguments can be extended with respect to the mean of patch and positional embeddings because LN applies normalization using the mean and standard deviation. Hence, we seek both mean and variance consistency simultaneously. Unless specified otherwise, we mainly examine variance when mean consistency holds (See the Appendix \ref{sec:empi}, \ref{sec:meanpe}).

\paragraph{Properties of Variance} By definition, variance provides a suitably scaled result with respect to its size. Because the variance of a tensor or matrix will be computed on its flattened version, we consider them as vectors. For a vector $\bv$, each element $v$ is sampled from the same distribution. By definition, variance is invariant to shuffling the order of elements in a vector. Because variance is scaled with respect to the sample size, duplicating the vector multiple times yields the same variance: $\Var[(\bv; \cdots; \bv)] = \Var[\bv]$. This property can be extended to the concatenation of different vectors (See the Appendix \ref{sec:proof}). Considering the practical scenario of vision tasks where the vector corresponds to a large feature map, we assume that the size of the vector is sufficiently large. The large number of samples assures that cropping a vector to use a subset of the vector yields approximately the same variance: $\Var[\Crop(\bv)] \approx \Var[\bv]$. Considering the practical scenario, we assume that the vector is not a constant vector.

\subsection{Variance Shift Due to Upsampling}
For an input image with a spatial size of $N_h \times N_w$, using a patch size of $N_p$ yields a patch embedding with a spatial size of $(N_h/N_p) \times (N_w/N_p)$, and the additive positional embedding has the same spatial size of $(N_h/N_p) \times (N_w/N_p)$. Although the common practice for training ViT is to use a fixed size of training image, such as $224 \times 224$, the image size may differ during the test phase. The use of an arbitrary size for the test image precludes the addition of positional embedding that has a fixed size. To address this issue, the original ViT study \citep{DBLP:conf/iclr/DosovitskiyB0WZ21} proposed upsampling the positional embedding, where bicubic interpolation is used with an upsampling rate determined by the size of the input image. Though upsampling the positional embedding might seem like a valid approach to cope with an arbitrary size of image, we claim that upsampling operations can affect variance, which causes the vulnerability of variance shift:
\begin{theorem}
	\label{thm:vardecrease}
	Upsampling yields $\E[\UP(\bv)] = \E[\bv]$ and $\Var[\UP(\bv)] = \Var[\bv]$ for duplication-type upsampling but not for interpolation-type upsampling.
\end{theorem}
For example, bicubic or bilinear upsampling corresponds to the interpolation-type, which cannot conserve both mean and variance simultaneously after its upsampling. Duplication-type upsampling indicates repeating exact elements in the vector, where the nearest neighbor upsampling corresponds. Indeed, nearest neighbor upsampling is equivalent to duplicating a vector and shuffling elements into the correct order, which conserves variance. This statement is stronger than Theorem 3.2 of Reference \citep{DBLP:journals/corr/abs-2402-01149}.

\begin{table}[t!]
	\caption{We empirically measured the variance ratio for different upsampling methods. Bicubic and bilinear upsamplings exhibit $k<1$, whereas nearest neighbor upsampling yields $k=1$.}
	\label{tab:ratio}
	\centering
	\begin{tabular}{l|ccc}
		\toprule
		\textbf{Measurement} & \textbf{Bicubic} & \textbf{Bilinear} & \textbf{Nearest Neighbor} \\
		\midrule
		$k_{2D}$             & 0.7295           & 0.3927            & 1.0000                    \\
		$k_{1D}$             & 0.8541           & 0.6267            & 1.0000                    \\
		\midrule
		$1/\sqrt{k_{2D}}$    & 1.1708           & 1.5957            & 1.0000                    \\
		$1/\sqrt{k_{1D}}$    & 1.0820           & 1.2632            & 1.0000                    \\
		\bottomrule
	\end{tabular}
\end{table}

Here, we rewrite our theorem as $\Var[\UP(\bv)] = k \Var[\bv]$ where $k=1$ for duplication-type upsampling and $k \neq 1$ for interpolation-type upsampling. The variance ratio $k$ is a constant that depends on the specific interpolation method (\tabref{tab:ratio}). The upsampling rate to adjust spatial size is considered to be a real number larger than one, and its choice does not affect the variance ratio $k$. This statement applies to arbitrary dimensional upsampling: Considering 1D upsampling that yields scaling by $k_{1D}$, 2D upsampling can be described as a repetition of 1D upsampling for two sides, and thus we obtain $k_{2D} = k_{1D}^2 \neq 1$ for the interpolation-type, and similarly $k_{2D} = k_{1D}^2 = 1$ for the duplication-type. Note that variance itself provides a suitably scaled result with respect to its size, which ensures that different variance is not caused by the increased size from upsampling. See the Appendix \ref{sec:proof} for a detailed proof and discussion. We are not saying that nearest neighbor upsampling is superior; rather, our claim is that as long as we follow the current practice of applying bicubic upsampling to positional embedding to match the size, variance exhibits inconsistency. Now, we investigate practical vision tasks where the upsampling operation incurs a variance shift in positional embedding.

\paragraph{Image Classification Scenario}
Since the introduction of VGGNet \citep{DBLP:journals/corr/SimonyanZ14a}, for training image classification tasks such as ImageNet, \texttt{RandomResizeCrop} has been widely deployed. This operation upsamples the training image to a slightly larger size and crops it to a fixed size, such as $224 \times 224$ pixels. During the test phase, arbitrary image size $N_h \times N_w$ is allowed, but similarly, using an inference crop percentage $p$, a resize operation into $N_h/p \times N_w/p$ and a center crop operation into $N_h \times N_w$ are applied in a row. For example, using inference crop percentage 0.9 and a test image size $224 \times 224$, it applies a resize operation to $248 \times 248$ and a center crop operation to $224 \times 224$.

\begin{figure}[t!]
	\centering
    \includegraphics[width=0.95\textwidth]{"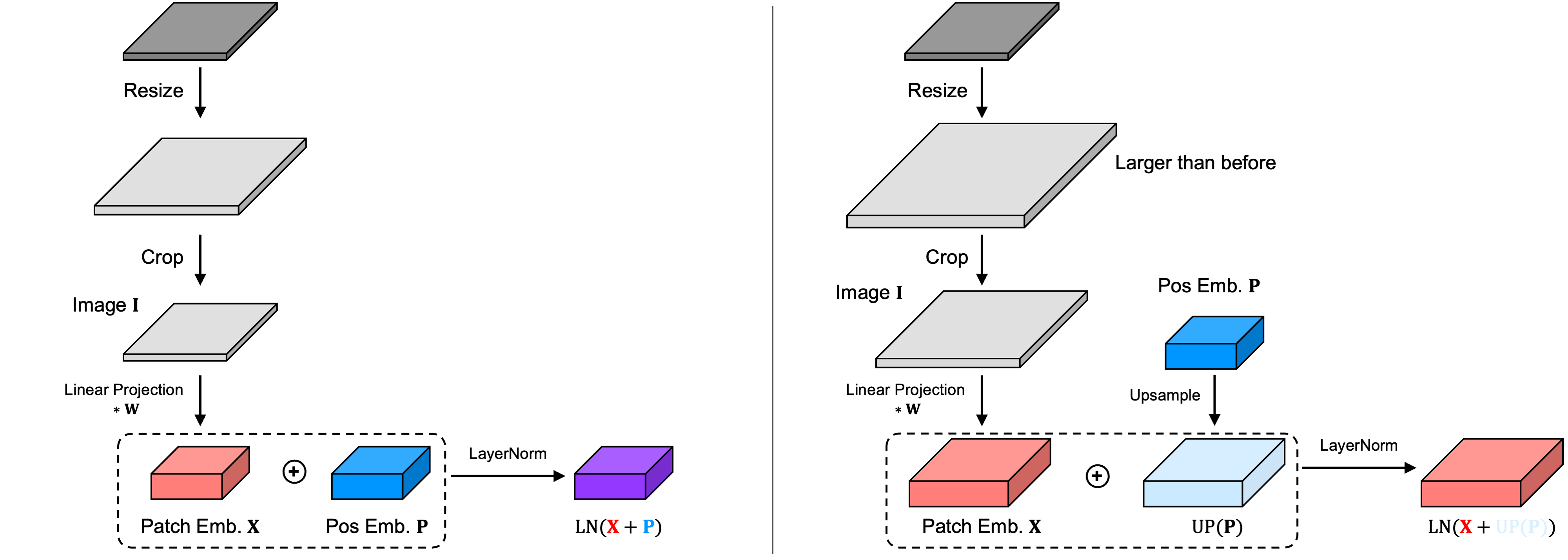"}
	\caption{Positional embedding is upsampled depending on the size of patch embedding, which causes inconsistent variance of positional embedding}
	\label{fig:resizecrop}
\end{figure}

Thus, the image is subjected to resize and crop operations, which decrease variance. However, whether to upsample the positional embedding depends on the size of the image. When positional embedding is not subjected to upsampling during the test phase, no variance shift occurs. The problem arises when using an arbitrary size of a test image, which requires upsampling of both the input image and positional embedding. This scenario breaks the consistency of the variance ratio (\figref{fig:resizecrop}). Specifically, we obtain
\begin{align}
	\Var[\bItr]               & = \Var[\Crop(\UP(\bI))] = k \Var[\bI], \label{eq:1}        \\
	\Var[\bPtr]               & = \Var[\bP] = \Var[\bP],               \label{eq:2}        \\
	\Var[\bIte]               & = \Var[\Crop(\UP(\bI))] = k \Var[\bI],        \label{eq:3} \\
	\Var[\bPte]               & = \Var[\UP(\bP)] = k \Var[\bP],        \label{eq:4}        \\
	\Var[\bItr] / \Var[\bPtr] & = k \Var[\bI] / \Var[\bP],             \label{eq:5}        \\
	\Var[\bIte] / \Var[\bPte] & = \Var[\bI] / \Var[\bP]. \label{eq:6}
\end{align}
Therefore, if positional embedding is subjected to upsampling, its variance varies, whereas the variance of the image---equivalently variance of patch embedding---remains the same. In this scenario, the variance of positional embedding during the test phase is $k$ times of that during the training phase, which we call the variance shift.

To remove the variance shift, we claim that the variance of positional embedding should be calibrated. One may attempt to match variances by replacing bicubic upsampling with nearest neighbor upsampling, which conserves variance. However, we observed that nearest neighbor upsampling produces poor performance because it provides unnatural interpolation that is unsuitable for visual and perceptual tasks. Considering this observation, we should employ natural upsampling such as bicubic upsampling while its variance shift should be removed. Here, we found that simply rescaling the upsampled positional embedding by $1/\sqrt{k}$ during the test phase works suitably in practice. In other words, to remove variance shift in the above scenario, we propose amplifying positional embedding during the test phase by $\bPte = \UP(\bP)/\sqrt{k}$, whose variance is $\Var[\bP]$, ensuring no variance shift.

The proposed rescaling is compatible with the current source code of ViTs and can be implemented by inserting few lines of code during the test phase; moreover, the original ViT model can be used without any retraining. The value $1/\sqrt{k}$ for each upsampling is summarized in \tabref{tab:ratio}, which does not need hyperparameter tuning. When the rescaled positional embedding is pre-computed and saved in advance, it does not impose additional computational cost in the main inference, which is actually a free gain beyond the trade-off between computational cost and performance.

\paragraph{Semantic Segmentation Scenario}
We additionally describe the variance shift for a downstream task, specifically in the semantic segmentation scenario. Semantic segmentation is one of the major fields in computer vision and refers to the task of generating a semantic mask that classifies each pixel in an image into a specific category. A semantic segmentation network with an encoder-decoder architecture extracts segmentation output that has the same size as the input image, which enables the use of an arbitrary size of the input image. For semantic segmentation on the ADE20K dataset \citep{DBLP:conf/cvpr/ZhouZPFB017} as an example, the common pipeline for data augmentation includes \texttt{RandomResize} with a scale (2048, 512) and \texttt{RandomCrop} with a size $512 \times 512$ during the training phase, whereas it applies \texttt{Resize} with a scale (2048, 512) without cropping during the test phase. The resize scale (2048, 512) indicates an upsampling where the maximum edge is no longer than 2048 and the shorter edge is no longer than 512. This behavior yields 1D or 2D upsampling depending on the dataset.

We interpret this practice as follows. Upsampling is applied to images during both the training and test phases, whereas cropping is only applied during the training phase, yielding $\Var[\bIte] = \Var[\UP(\bI)] = k \Var[\bI]$. In this case, the input image to ViT has a larger spatial size during the test phase compared with the training phase. To cope with the larger spatial size of the image during the test phase, bicubic upsampling is applied to positional embedding only during the test phase. Therefore, semantic segmentation with ViT yields the same equations as \cref{eq:1,eq:2,eq:3,eq:4,eq:5,eq:6} except for the minor difference \cref{eq:3}, and similarly, we propose applying the $1/\sqrt{k}$ rescaling to positional embedding.

\subsection{Variance Shift Due to Mixup}
\begin{figure}[t!]
	\centering
    \includegraphics[width=0.90\textwidth]{"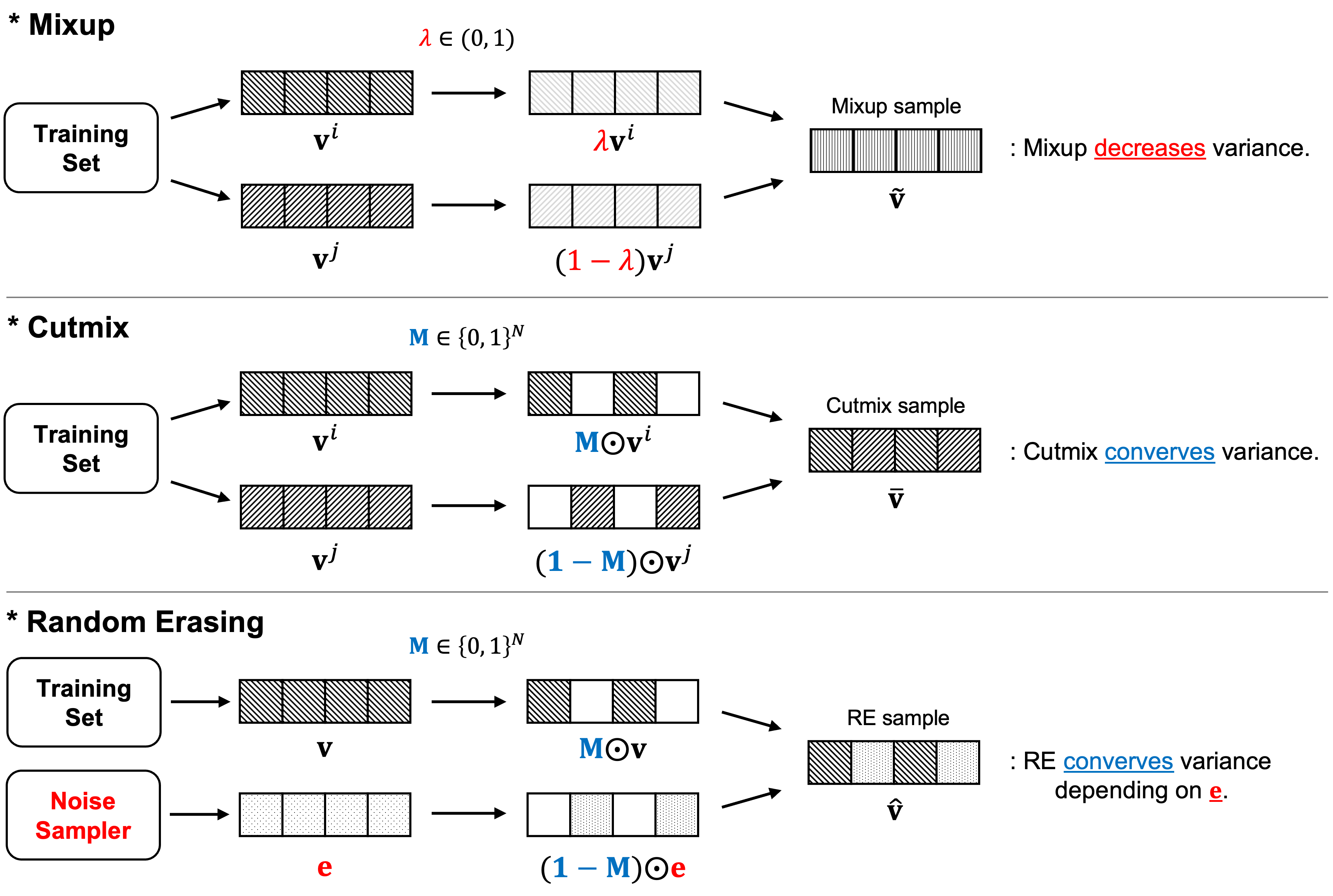"}
	\caption{Illustration of Mixup, Cutmix, and random erasing. The level of darkness indicates variance.}
	\label{fig:dataaug}
\end{figure}

For image classification tasks, several advanced data augmentations, such as Mixup \citep{DBLP:conf/iclr/ZhangCDL18} and Cutmix \citep{DBLP:conf/iccv/YunHCOYC19}, have been proposed. They proposed using a combination of two images for training, which substantially boosted the performance of CNNs. These data augmentations have also been adopted for training ViTs. Indeed, several studies on ViT have revealed their training recipes, including data augmentations and exact hyperparameters \citep{DBLP:conf/icml/TouvronCDMSJ21,DBLP:conf/eccv/TouvronCJ22,DBLP:conf/iclr/Bao0PW22,DBLP:conf/cvpr/HeCXLDG22,DBLP:conf/iccv/LiuL00W0LG21,DBLP:conf/cvpr/Liu0LYXWN000WG22,DBLP:conf/iccv/ChenFP21}. In their training recipes, Cutmix has been steadily preferred. Mixup, however, has been used with a Mixup ratio set to 0.8 to have a slight effect while avoiding a Mixup ratio of 0.5, which corresponds to its maximum usage. Furthermore, the study of EVA \citep{DBLP:conf/cvpr/FangWXSWW0WC23} reported that they turned off Mixup in their training recipe. These practices are explainable through our analysis: We claim that Mixup causes variance shift, whereas Cutmix conserves variance (\figref{fig:dataaug}).

Let $\bv^i$ and $\bv^j$ be vectors sampled from the training distribution, where $\E[\bv^i] = \E[\bv^j]$ and $\Var[\bv^i] = \Var[\bv^j]$. Mixup proposes training a neural network using a Mixup sample $\tbv = \lambda \bv^i + (1-\lambda) \bv^j$ with a Mixup ratio $\lambda \in (0, 1)$. For the Mixup scheme, we obtain the following property:
\begin{proposition}
	Mixup decreases variance: $\Var[\tbv] < \Var[\bv]$.
\end{proposition}
The proof is straightforward because $\lambda^2 + (1-\lambda)^2 \neq 1$ unless $\lambda=0$ or $\lambda=1$, \ie, as long as it is a valid Mixup sample. The inconsistent variance can be observed for other $\lambda$ beyond $(0, 1)$ and for other variants of Mixup (See the Appendix \ref{sec:varmixup}).

In contrast, Cutmix combines two samples using a binary mask $\bM$ where each element is either zero or one. Cutmix proposes training a neural network using a Cutmix sample $\bbv = \bM \odot \bv^i + (\bone-\bM) \odot \bv^j$, where $\odot$ indicates element-wise multiplication. For the Cutmix scheme, we obtain the following property:
\begin{proposition}
	Cutmix conserves variance: $\Var[\bbv] = \Var[\bv]$.
\end{proposition}
Because variance is invariant to the shuffling order of a vector, without loss of generality, we may place mask elements of one at the front and zero at the back. This permutation enables us to describe the Cutmix sample as a concatenation of cropped vectors: $[\bv^{i,\prime}_{:\sum{\bM}}; \bv^{j,\prime}_{\sum{\bM}+1:}]$. Because shuffling, cropping, and concatenation conserve variance, the Cutmix sample exhibits the same variance as that of $\bv$. The same goes for the mean.

Furthermore, numerous variants of Cutmix have been proposed, which have modified the sampling of binary masks $\bM$ for a more natural combination of images \citep{DBLP:conf/icml/KimCS20,DBLP:conf/iclr/UddinMSCB21,DBLP:conf/icassp/WalawalkarSLS20,DBLP:conf/aaai/HuangWT21}. We find that all these Cutmix variants still follow the masked combination scheme $\bbv = \bM \odot \bv^i + (\bone-\bM) \odot \bv^j$, which conserves variance. In summary, Cutmix and its variants conserve the variance of the input image, which prevents and is safe from the variance shift in positional embedding. However, we claim that the use of Mixup should be reconsidered for training ViT because it causes variance shifts. Note that we are not saying Mixup is wrong; rather, our claim is that Mixup has a side effect of variance shift, which may outweigh the advantage of training combined samples.

\subsection{Variance Shift Due to Random Erasing}
Random erasing \citep{DBLP:conf/aaai/Zhong0KL020}, also referred to as Cutout \citep{DBLP:journals/corr/abs-1708-04552}, is a widely adopted data augmentation technique. Random erasing randomly selects and drops a certain block from the input image during training. The original study of random erasing proposed several versions for erasing behavior, such as replacing the block with zero (\texttt{const} mode), replacing the block with a single random normal value (\texttt{rand} mode), and replacing the block with a per-pixel random normal value (\texttt{pixel} mode). Again, to investigate the variance shift in positional embedding, we examine the variance of the image after random erasing. Our findings are as follows:
\begin{proposition}
	Random erasing conserves variance if we configure it to \texttt{pixel} mode and adopt the correct mean-std normalization using dataset statistics.
\end{proposition}

In fact, replacing a certain block of the input image is similar to the Cutmix scheme. For a vector $\bv$, random erasing selects a binary mask $\bM$ where each element is either zero or one and replaces the corresponding block with that of the erasing vector $\be$, which yields a random erasing sample $\hbv = \bM \odot \bv + (\bone-\bM) \odot \be$. Similar to the Cutmix scheme, when investigating variance, we may permute the order of the vector to describe it as concatenation: $[\bv^{\prime}_{:\sum{\bM}}; \be^{\prime}_{\sum{\bM}+1:}]$. The property of the concatenated vector can be examined through Lemma \ref{lem:concat} in the Appendix \ref{sec:proof}, which says that the concatenated vector achieves $\E[\hbv] = \E[\bv]$ and $\Var[\hbv] = \Var[\bv]$ if $\E[\bv] = \E[\be]$ and $\Var[\bv] = \Var[\be]$. This property holds for an arbitrary drop probability of random erasing.

Here, the \texttt{const} mode provides an erasing vector as a zero vector $\be = \bzero$, the \texttt{rand} mode samples a single value $e_i \sim \N(0, 1)$ shared across all elements, and the \texttt{pixel} mode independently samples each element $e_i \sim \N(0, 1)$. Thus, in \texttt{const} and \texttt{rand} modes, the erasing vector $\be$ comprises a single value, which exhibits $\Var[\be] = 0$ and thereby causes $\Var[\hbv] \neq \Var[\bv]$. Only the \texttt{pixel} mode has different elements that are sampled from $\N(0, 1)$, which leads to $\Var[\be] = 1$. Therefore, to maintain the consistency of mean and variance after random erasing, we should ensure the \texttt{pixel} mode, $\E[\bv] = 0$, and $\Var[\bv] = 1$.

Fortunately, the conditions $\E[\bv] = 0$ and $\Var[\bv] = 1$ tend to hold for standard training recipes because applying global mean-std normalization using dataset statistics has been standard practice. When training ViT on the ImageNet dataset, researchers have used mean-std normalization with pre-computed ImageNet statistics of per-channel mean $(0.485, 0.456, 0.406)$ and standard deviation (std) $(0.229, 0.224, 0.225)$. These statistics are referred to as default-mean-std, which guarantees $\E[\bv] = 0$ and $\Var[\bv] = 1$ for the input images and ensures consistent variance after random erasing with \texttt{pixel} mode.

However, several exceptions violate this condition. A prime example is so-called inception-mean-std \citep{DBLP:conf/cvpr/SzegedyLJSRAEVR15,DBLP:journals/tmlr/SteinerKZWUB22}, which applies mean-std normalization using per-channel mean $(0.5, 0.5, 0.5)$ and std $(0.5, 0.5, 0.5)$. Adopting these values causes $\E[\bv] \neq 0$ and $\Var[\bv] \neq 1$ and thereby inconsistent mean and variance after random erasing, even when using the \texttt{pixel} mode. In fact, the original implementation of ViT \citep{DBLP:conf/iclr/DosovitskiyB0WZ21} preferred inception-mean-std. Note that the original study of ViT used a much larger dataset of JFT-300M with few data augmentations that excluded random erasing, which is safe from variance shifts due to random erasing. However, other follow-up studies have trained ViT and its variants using the ImageNet dataset with rich data augmentations including random erasing, which is vulnerable to the inception-mean-std. Because using random erasing is a standard practice for training ViT, we claim that we should configure it as \texttt{pixel} mode and simultaneously adopt correct mean-std normalization using the default-mean-std, avoiding the inception-mean-std.

We find that the use of inception-mean-std in the original implementation of ViT has influenced other follow-up studies on variants of ViT. For example, the original study of BEiT \citep{DBLP:conf/iclr/Bao0PW22} used inception-mean-std, whereas BEiT v2 \citep{DBLP:journals/corr/abs-2208-06366} opted for the default-mean-std using ImageNet statistics. For certain models of MaxViT \citep{DBLP:conf/eccv/TuTZYMBL22}, inception-mean-std was used, whereas for other models, default-mean-std was adopted. Furthermore, the pretrained ViTs provided by MIIL \citep{DBLP:conf/nips/RidnikBNZ21} simply used per-channel mean $(0, 0, 0)$ and std $(1, 1, 1)$. Considering these practices where the default-mean-std is unused but random erasing is used in the training of ViT, we claim that a variance shift arises, and therefore we should opt for the correct mode of random erasing as well as the correct mean-std normalization using dataset statistics.

\section{Experiments}
\subsection{Rescaling Positional Embedding}
\paragraph{ImageNet Evaluation} We examine the effect of rescaling positional embedding when using upsampled images during the test phase. We target ViTs pretrained ImageNet, provided by AugReg \citep{DBLP:journals/tmlr/SteinerKZWUB22} that pretrained ViTs on ImageNet-21K and fine-tuned on ImageNet-1K with $224 \times 224$ size. We examined ViT-\{S, B, L\} with a patch size of 16, where S, B, and L stand for small, base, and large models, respectively. For these models, we examined the test size across $224 \times 224$ to $864 \times 864$. For each test size, we first evaluated the top-1 accuracy on the validation set of ImageNet using the existing practice of upsampling positional embedding $\UP(\bP)$, which corresponds to the baseline performance. Subsequently, we performed the same evaluation procedures with rescaled positional embedding $\UP(\bP)/\sqrt{k}$, which corresponds to the proposed method to prevent variance shift.

\begin{table}[t!]
	\caption{Top-1 accuracy (\%) for ImageNet with respect to various spatial sizes. Baseline indicates the existing practice of using $\UP(\bP)$, and ours indicates calibrating the variance using $\UP(\bP)/\sqrt{k}$.}
	\label{tab:in}
	\centering
	\resizebox{\textwidth}{!}{% <------ Don't forget this %
		\begin{tabular}{l|ccc|ccc|ccc}
			\toprule
			              & \multicolumn{3}{c|}{\textbf{ViT-S/16 $\mathbf{224^2}$}} & \multicolumn{3}{c|}{\textbf{ViT-B/16 $\mathbf{224^2}$}} & \multicolumn{3}{c}{\textbf{ViT-L/16 $\mathbf{224^2}$}}                                                                                                         \\
			\textbf{Size} & \textbf{Baseline}                                       & \textbf{Ours}                                           & \textbf{Diff}                                          & \textbf{Baseline} & \textbf{Ours} & \textbf{Diff} & \textbf{Baseline} & \textbf{Ours} & \textbf{Diff} \\
			\midrule
			\rowcolor[rgb]{0.8, 0.8, 0.8}
			$224^2$       & 81.386                                                  & 81.296                                                  & -0.090                                                 & 84.528            & 84.344        & -0.184        & 85.834            & 85.570        & -0.264        \\
			\midrule
			$288^2$       & 82.024                                                  & 82.044                                                  & +0.020                                                 & 84.810            & 84.846        & +0.036        & 86.288            & 86.382        & +0.094        \\
			$352^2$       & 81.438                                                  & 81.470                                                  & +0.032                                                 & 84.204            & 84.406        & +0.202        & 85.762            & 85.788        & +0.026        \\
			$416^2$       & 80.008                                                  & 80.320                                                  & +0.312                                                 & 83.056            & 83.320        & +0.264        & 84.944            & 85.026        & +0.082        \\
			$480^2$       & 77.928                                                  & 78.472                                                  & +0.544                                                 & 81.538            & 82.100        & +0.562        & 84.034            & 84.158        & +0.124        \\
			$544^2$       & 75.424                                                  & 76.296                                                  & +0.872                                                 & 79.374            & 80.422        & +1.048        & 82.868            & 82.924        & +0.056        \\
			$608^2$       & 72.384                                                  & 73.722                                                  & +1.338                                                 & 76.744            & 78.384        & +1.640        & 81.492            & 81.590        & +0.098        \\
			$672^2$       & 68.964                                                  & 70.882                                                  & +1.918                                                 & 73.560            & 76.048        & +2.488        & 79.982            & 80.106        & +0.124        \\
			$736^2$       & 65.014                                                  & 67.704                                                  & +2.690                                                 & 69.832            & 73.324        & +3.492        & 78.494            & 78.654        & +0.160        \\
			$800^2$       & 60.954                                                  & 64.196                                                  & +3.242                                                 & 65.674            & 70.524        & +4.850        & 76.800            & 77.096        & +0.296        \\
			$864^2$       & 56.528                                                  & 60.710                                                  & +4.182                                                 & 61.330            & 67.280        & +5.950        & 75.282            & 75.656        & +0.374        \\
			\bottomrule
		\end{tabular}
	}
\end{table}

\tabref{tab:in} summarizes the results. Firstly, upsampling to larger sizes incurs a distribution shift of facing different images, which yields decreased top-1 accuracy. However, we discovered that the decreased performance is not only caused by different images but also by the variance shift of positional embedding: Rescaling the positional embedding removed the hidden factor of degraded performance and recovered the top-1 accuracy to a certain degree. Indeed, our analysis says that when using a test size larger than $224 \times 224$, we should rescale positional embedding. The results were exactly in agreement with our claim: Rescaling the positional embedding improved top-1 accuracy when using a test size larger than $224 \times 224$ but degraded performance when using a test size of $224 \times 224$, which corresponds to the case that does not require upsampling the positional embedding.

\begin{table}[t!]
	\caption{Results on semantic segmentation before and after applying $1/\sqrt{k}$ rescaling to positional embedding. The proposed rescaling consistently improved the performance.}
	\label{tab:ss}
	\centering
	\begin{tabular}{ll|rrr}
		\toprule
		\textbf{Dataset}            & \textbf{Index (\%)} & $\UP(\bP)$ & $\UP(\bP)/\sqrt{k}$ & \textbf{Diff} \\
		\midrule
		\multirow{3}{*}{ADE20K}     & aAcc                & 80.73      & 80.74               & +0.01         \\
		                            & mIoU                & 43.16      & 43.27               & +0.11         \\
		                            & mAcc                & 54.28      & 54.36               & +0.08         \\
		\midrule
		\multirow{3}{*}{LoveDA}     & aAcc                & 65.58      & 66.46               & +0.88         \\
		                            & mIoU                & 47.78      & 48.37               & +0.59         \\
		                            & mAcc                & 62.00      & 62.20               & +0.20         \\
		\midrule
		\multirow{3}{*}{Cityscapes} & aAcc                & 96.02      & 96.04               & +0.02         \\
		                            & mIoU                & 77.20      & 77.31               & +0.11         \\
		                            & mAcc                & 84.27      & 84.40               & +0.13         \\
		\bottomrule
	\end{tabular}
\end{table}

\paragraph{Semantic Segmentation} Now we examine rescaling the positional embedding targeting semantic segmentation tasks. The ADE20K dataset \citep{DBLP:conf/cvpr/ZhouZPFB017} contains scene-centric images along with the corresponding segmentation labels. A crop size of $512 \times 512$ pixels was used, which was obtained after applying mean-std normalization and a random resize operation using a resize scale of (2048, 512) pixels with a ratio range of 0.5 to 2.0. Furthermore, a random flipping with a probability of 0.5 and the photometric distortions were applied. The objective was to classify each pixel into one of the 150 categories and train the segmentation network using the pixel-wise cross-entropy loss. The same goes for the LoveDA dataset \citep{DBLP:conf/nips/WangZMLZ21} with 7 categories and the Cityscapes dataset \citep{DBLP:conf/cvpr/CordtsORREBFRS16} with 19 categories, except for using a crop size of $1024 \times 1024$ pixels for the Cityscapes dataset. Note that the resize operation applies 1D upsampling or 2D upsampling depending on the dataset; 1D upsampling is applied for the ADE20K and Cityscapes datasets, whereas 2D upsampling is applied for the LoveDA dataset. The rescaling values $1/\sqrt{k}$ were referred to in \tabref{tab:ratio}.

We used DeiT-S/16 \citep{DBLP:conf/icml/TouvronCDMSJ21} pretrained on ImageNet and UPerNet \citep{DBLP:conf/eccv/XiaoLZJS18} with a multi-level neck. For the ADE20K dataset, pretrained segmentation networks were downloaded; for the LoveDA and Cityscapes datasets, we trained segmentation networks by ourselves. To follow common practice for semantic segmentation, training recipes from \texttt{MMSegmentation} were employed \citep{mmseg2020}. For training, AdamW optimizer with weight decay $10^{-2}$, betas $\beta_1 = 0.9$, $\beta_2 = 0.999$, and learning rate $6 \times 10^{-5}$ with polynomial decay of the 160K scheduler after linear warmup were used. The training was performed on a 4$\times$A100 GPU machine.

We measured three indices commonly used in semantic segmentation---all pixel accuracy (aAcc), mean accuracy of each class (mAcc), and mean intersection over union (mIoU) (\tabref{tab:ss}). The baseline corresponds to using $\UP(\bP)$, whereas the proposed method corresponds to $\UP(\bP)/\sqrt{k}$. We observed that for all datasets and indices, rescaling the positional embedding consistently improved the segmentation performance.

\subsection{Effect of Mixup and Cutmix}
\paragraph{ImageNet Training} We examine the effect of Mixup and Cutmix in training ViTs. In contrast to previous experiments, we now train ViTs ourselves. The ImageNet dataset contains 1.28M images for 1,000 classes. For the image classification experiments with ImageNet, we used the pytorch-image-models library, also known as \texttt{timm} \citep{rw2019timm}. We referred to the hyperparameter recipe described in the official documentation and the recipe from DeiT. For training, AdamW optimizer \citep{DBLP:conf/iclr/LoshchilovH19} with learning rate $5 \times 10^{-4}$, epochs 300, warm-up learning rate $10^{-6}$, cosine annealing schedule \citep{DBLP:conf/iclr/LoshchilovH17}, weight decay 0.05, label smoothing \citep{DBLP:conf/cvpr/SzegedyVISW16} 0.1, RandAugment \citep{DBLP:conf/nips/CubukZS020} of magnitude 9 and noise-std 0.5 with increased severity (rand-m9-mstd0.5-inc1), stochastic depth \citep{DBLP:conf/eccv/HuangSLSW16} 0.1, mini-batch size 288 per GPU, Exponential Moving Average of model weights with decay factor 0.99996, and image resolution $224 \times 224$ were used. The training was performed on an 8$\times$A100 GPU machine, which requires from two to three days per training.

\begin{table}[t!]
	\begin{minipage}{.50\linewidth}
		\caption{Top-1 accuracy (\%) for ImageNet. Cutmix improves performance, whereas Mixup does not.}
		\label{tab:mucm}
		\centering
		\resizebox{0.99\textwidth}{!}{% <------ Don't forget this %
			\begin{tabular}{ll|rr}
				\toprule
				\textbf{Mean-Std}        & \textbf{Data Augmentation} & \multicolumn{1}{l}{\textbf{ViT-S/16}} & \multicolumn{1}{l}{\textbf{ViT-B/16}} \\
				\midrule
				\multirow{4}{*}{Default} & Mixup 0.0, Cutmix 0.0      & 74.801                                & 74.741                                \\
				                         & Mixup 0.0, Cutmix 1.0      & \textbf{79.540}                       & \textbf{80.240}                       \\
				                         & Mixup 0.2, Cutmix 1.0      & 78.069                                & 79.448                                \\
				                         & Mixup 0.4, Cutmix 1.0      & 78.128                                & 79.248                                \\
				\bottomrule
			\end{tabular}
		}
	\end{minipage}%
	\begin{minipage}{.50\linewidth}
		\centering
		\caption{Top-1 accuracy (\%) for ImageNet. When using random erasing (RE), its \texttt{pixel} mode and default-mean-std configuration are required.}
		\label{tab:remode}
		\resizebox{0.99\textwidth}{!}{% <------ Don't forget this %
			\begin{tabular}{ll|rr}
				\toprule
				\textbf{Mean-Std}          & \textbf{Data Augmentation}       & \multicolumn{1}{l}{\textbf{ViT-S/16}} & \multicolumn{1}{l}{\textbf{ViT-B/16}} \\
				\midrule
				\multirow{3}{*}{Inception} & RE 0.00 with \texttt{pixel} mode & \textbf{79.426}                       & \textbf{80.134}                       \\
				                           & RE 0.25 with \texttt{pixel} mode & 79.126                                & 79.894                                \\
				                           & RE 0.50 with \texttt{pixel} mode & 78.726                                & 79.794                                \\
				\midrule
                \multirow{3}{*}{Default}   & RE 0.25 with \texttt{pixel} mode & \textbf{79.158}                       & \textbf{80.415}                       \\
				                           & RE 0.25 with \texttt{rand} mode  & 79.009                                & 80.398                                \\
				                           & RE 0.25 with \texttt{const} mode & 78.909                                & 80.298                                \\
				\bottomrule
			\end{tabular}
		}
	\end{minipage}
\end{table}

\tabref{tab:mucm} summarizes the results. Comparing the results without and with Cutmix, we find that Cutmix is significantly beneficial to performance. However, when additionally applying Mixup, the top-1 accuracy rather decreased, which indicates that the side effect of variance shift outweighs the advantage of training combined samples of Mixup. These results are in agreement with our claim: Cutmix is suitable for training ViT, whereas Mixup is not owing to variance shift.

\subsection{Configurations of Random Erasing}
Now, we examine the effect of configuration on random erasing using different probabilities and modes. We used the same recipe as before, but without Mixup and with Cutmix.

\tabref{tab:remode} summarizes the result. When using inception-mean-std, applying random erasing rather decreased the top-1 accuracy, and the recipes without random erasing worked better. In contrast, when using default-mean-std, the best result was found for the \texttt{pixel} mode, whereas \texttt{rand} and \texttt{const} modes yielded slightly lower performance. All these results validate our claim: When applying random erasing, we should choose the default-mean-std and \texttt{pixel} mode.

\section{Conclusion}
This study reported the variance shift in the positional embedding of ViTs caused by data augmentations. We discussed four data augmentations: random resize crop, Mixup, Cutmix, and random erasing. Firstly, we showed that common upsampling techniques lead to inconsistent variance and described a resultant variance shift in positional embedding. Furthermore, we identified that Mixup incurs variance shifts, whereas Cutmix is safe from this issue. We further analyzed the detailed behavior of random erasing to reveal its correct configuration for variance consistency---\texttt{pixel} mode with mean-std normalization using dataset statistics. The proposed methods were validated through various experiments on ViTs, where removing the variance shift in positional embedding consistently improved the performance of ViTs.

\bibliography{mybib}
\bibliographystyle{plain}

%%% main - ref - appendix - checklist

\appendix

\section{Proof of Theorem \ref{thm:vardecrease}}
\label{sec:proof}

\paragraph{Notation} We consider a scenario where a vector $(x_1, x_2, \cdots, x_n)$ is subjected to 1D upsampling, which can be extended to 2D upsampling. We consider a function $f$ that interpolates each interval between $x_i$ and $x_{i+1}$ that corresponds to coordinates $p_i$ and $p_{i+1}$. We reflect common upsampling behavior in PyTorch as much as possible. For inner intervals, the property of the function $f$ is determined by the interpolation method, such as bilinear, bicubic, and nearest neighbor upsamplings. For outer intervals, we obtain $f(p)=x_n$ for the right outer interval and $f(p)=x_1$ for the left outer interval. The upsampling can be interpreted as sampling new points using the function $f$ from another coordinate. We consider one arbitrary point $q_i$ for each interval between $p_i$ and $p_{i+1}$ and denote the sampled data as $y_i$ that corresponds to the coordinate $q_i$.

\begin{figure}[h!]
	\centering
	\includegraphics[width=1.0\textwidth]{"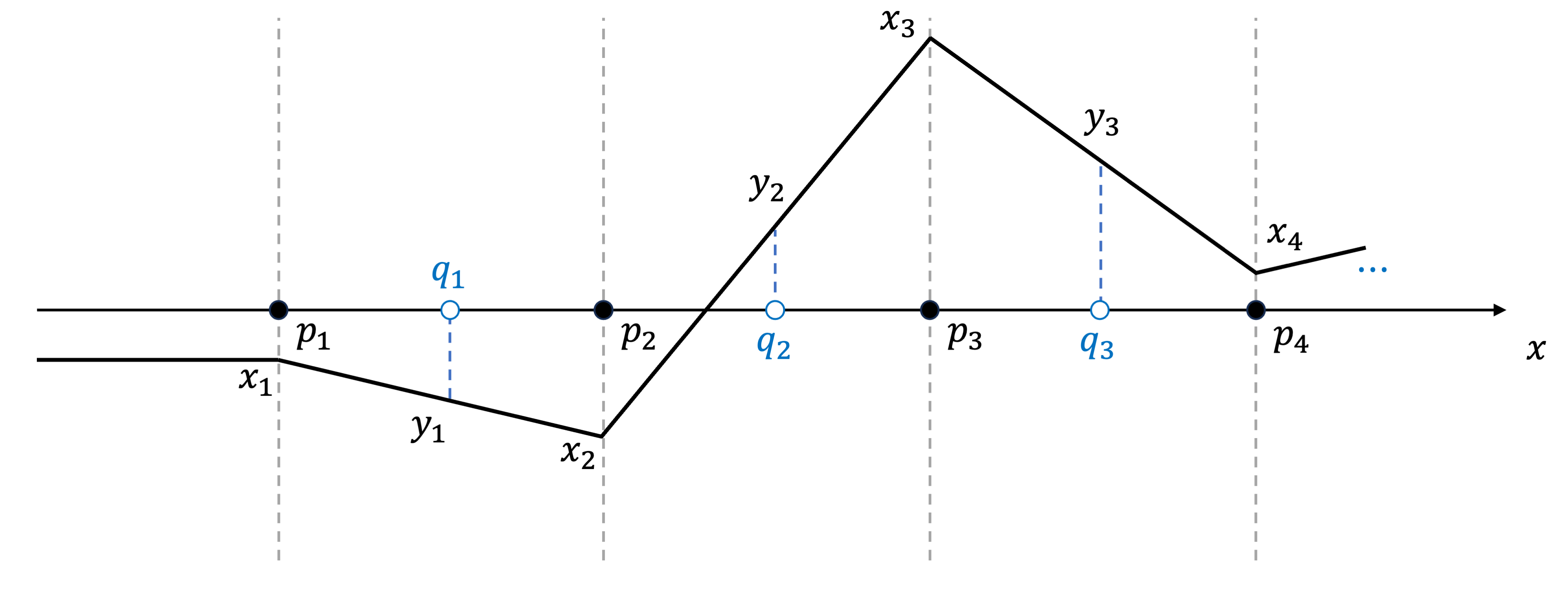"}
	\caption{Illustration of the coordinate and interpolation}
	\label{fig:new_coordinate}
\end{figure}

\paragraph{Auxiliary Property} Additionally, we use the following property of mean and variance:
\begin{lemma} \label{lem:concat}
	Consider two vectors $\bv^i$ and $\bv^j$. If $\E[\bv^i] = \E[\bv^j] = \mu$ and $\Var[\bv^i] = \Var[\bv^j] = \sigma^2$, then its concatenation exhibits the same mean and variance: $\E[(\bv^i ; \bv^j)] = \mu$ and $\Var[(\bv^i ; \bv^j)] = \sigma^2$.
\end{lemma}
This property arises from the definitions of mean and variance. Moreover, this holds true for an arbitrary size of the two vectors, say $n_i$ and $n_j$, because
\begin{align}
	\E[(\bv^i; \bv^j)]   & = \frac{1}{n_i + n_j} \{ (v^i_1 + \cdots + v^i_{n_i}) + (v^j_1 + \cdots + v^j_{n_j}) \}                                        \\
	                     & = \frac{1}{n_i + n_j} ( n_i \mu + n_j \mu )                                                                                    \\
	                     & = \mu,                                                                                                                         \\
	\Var[(\bv^i; \bv^j)] & = \frac{1}{n_i + n_j} [\{ (v^i_1-\mu)^2 + \cdots + (v^i_{n_i}-\mu)^2 \} + \{ (v^j_1-\mu)^2 + \cdots + (v^j_{n_j} - \mu)^2 \} ] \\
	                     & = \frac{1}{n_i + n_j} ( n_i \sigma^2 + n_j \sigma^2 )                                                                          \\
	                     & = \sigma^2.
\end{align}

\paragraph{Motivation} Using the above setup, we investigate the form of upsampling where both the mean and variance are conserved. Note that the mean and variance are invariant to shuffling. In our notation, upsampled data is written as $(x_1, y_1, x_2, y_2, \cdots, x_n, y_n)$. Permuting this vector yields $(x_1, x_2, \cdots, x_n, y_1, y_2, \cdots, y_n)$, which is a concatenation of $(x_1, x_2, \cdots, x_n)$ and $(y_1, y_2, \cdots, y_n)$. In other words, if the mean and variance of the new sample $(y_1, y_2, \cdots, y_n)$ are the same as those of the original one $(x_1, x_2, \cdots, x_n)$, then the upsampled vector conserves the mean and variance. From this observation, we examine the mean and variance of the new samples $(y_1, y_2, \cdots, y_n)$.

\paragraph{Main Proof} Now we analyze the conditions on the interpolation function $f$ that lead to consistency of mean and variance for the new samples $(y_1, y_2, \cdots, y_n)$. This consistency holds if the first and second momentum are conserved:
\begin{align}
	x_1 + x_2 + \cdots + x_n       & = y_1 + y_2 + \cdots + y_n, \label{eq:mean}      \\
	x_1^2 + x_2^2 + \cdots + x_n^2 & = y_1^2 + y_2^2 + \cdots + y_n^2. \label{eq:var}
\end{align}
In our setup, the new sample $y_i$ is determined by the interpolation function, which is affected by neighboring samples $x_i$ and $x_{i+1}$. For now, we attempt to express the new sample as a polynomial function of the two neighboring samples: $y_i = \sum_{p}{(a_{i,i,p}x_{i}^p + a_{i,i+1,p}x_{i+1}^p)}$. Although we started with a polynomial function of arbitrary order, owing to the mean consistency in \eqref{eq:mean}, all other terms are removed except for the first order, \ie, $y_i = a_{i,i,1} x_i + a_{i,i+1,1} x_{i+1}$. In other words, mean consistency requires a linear, first-order interpolation function.

Furthermore, these conditions should hold for arbitrary choices of upsampling subject $(x_1, x_2, \cdots, x_n)$. When comparing each coefficient in \eqref{eq:mean}, we obtain the following equations:
\begin{align}
	a_{1,1,1}             & = 1, \\
	a_{1,2,1} + a_{2,2,1} & = 1, \\
	a_{2,3,1} + a_{3,3,1} & = 1, \\
	\cdots
\end{align}
Similarly, from \eqref{eq:var}, we obtain
\begin{align}
	a_{1,1,1}^2               & = 1, \\
	a_{1,1,1} a_{1,2,1}       & = 0, \\
	a_{1,2,1}^2 + a_{2,2,1}^2 & = 1, \\
	\cdots
\end{align}
Solving the above equations yields $a_{i,i,1} = 1$ and $a_{i,i+1,1} = 0$ for all $i$. Thus, we obtain $y_i = x_i$ for all $i$. In other words, the interpolation function $f$ should behave as a copy of neighboring data to satisfy both mean and variance consistency. Others that act as smooth interpolations cannot be a solution. For example, bilinear interpolation is a linear interpolation, but it decreases variance, as described in Theorem 3.2 of Reference \citep{DBLP:journals/corr/abs-2402-01149}.

Although $y_i = x_i$ is a unique solution in the above-mentioned setup, other similar solutions exist for extended setups. This is because the variance is invariant to the shuffling order of the vector; copying $y_i = x_i$ and randomly shuffling it satisfies both mean and variance consistency. For example, a swapping operator $y_1 = x_2$ and $y_2 = x_1$ for two samples conserves the mean and variance. We collectively refer to these upsamplings as \textit{duplication-type}. Nearest neighbor upsampling is a special case of duplication-type upsampling because it can be represented as duplication and shuffling.

\section{On Variants of Mixup}
\label{sec:varmixup}
In the main text, we mentioned that Mixup and its variants result in inconsistent variance. For the Mixup sample $\tbv = \lambda \bv^i + (1-\lambda) \bv^j$, we obtain $\Var[\tbv] = (\lambda^2 + (1-\lambda)^2) \Var[\bv]$. Therefore, the Mixup sample exhibits inconsistent variance for an arbitrary Mixup ratio beyond $(0, 1)$ unless $\lambda=0$ or $\lambda=1$. However, $\lambda=0$ or $\lambda=1$ indicates that Mixup does not combine two samples.

One may attempt to design a variant of Mixup. Allowing different choices of two $\lambda$, we now try an extended form of Mixup sample $\dtbv = \lambda_i \bv^i + \lambda_j \bv^j$. For example, when choosing $\lambda_i = \lambda_j = 1/\sqrt{2}$, we obtain variance consistency $\Var[\dtbv] = \Var[\bv]$. In this Mixup, however, another side effect arises: It shows an inconsistent mean because $\E[\dtbv] = \sqrt{2} \E[\bv]$. Formally, we write
\begin{proposition}
	The extended form of the Mixup sample $\dtbv = \lambda_i \bv^i + \lambda_j \bv^j$ cannot conserve both mean and variance simultaneously.
\end{proposition}
In fact, when $\dtbv$ achieves mean consistency, we obtain $\lambda_j = 1-\lambda_i$, which leads to standard Mixup $\dtbv = \lambda \bv^i + (1-\lambda) \bv^j$, thereby breaking variance consistency.

\section{Dropout in Early Stage}
\label{sec:dropout}
Random erasing replaces certain blocks in an image with random values. Because the image is linearly projected into patch embedding, erasing certain patches is approximately equivalent to random erasing. This can be implemented by applying Dropout \citep{DBLP:journals/jmlr/SrivastavaHKSS14} to patch embedding. However, similar to random erasing, applying Dropout to patch embedding is still prone to variance shift.

Notably, recent studies on PatchDropout \citep{DBLP:conf/cvpr/Li0HFH23} suggested randomly discarding patches without replacements, which produces a subset of patches. The PatchDropout is a rather safe method to prevent variance shift because mean and variance are conserved when using a subset of a vector.

Another safe method is to apply Dropout to the sum of patch and positional embeddings. If both embeddings are subjected to Dropout, their relative variance ratio remains the same, which is free from variance shift. In other words, applying Dropout to one embedding yields a variance shift, whereas applying Dropout to both embeddings guarantees consistent variance. Indeed, the original studies of transformer \citep{DBLP:conf/nips/VaswaniSPUJGKP17} mentioned applying Dropout to the sum of two embeddings, which is safe from variance shift.

\section{Empirical Observation on Mean consistency}
\label{sec:empi}

In the main text, empirical measurements of variance ratios were reported. Additionally, we measured the mean ratio $K = \E[\UP(\bv)] / \E[\bv]$ for different upsampling methods (\tabref{tab:meanratio}). In fact, bicubic, bilinear, and nearest neighbor empirically showed mean consistency. The full source code is provided in the supplementary material.

\begin{table}[h!]
	\caption{Empirical results on mean ratio}
	\label{tab:meanratio}
	\centering
	\begin{tabular}{l|ccc}
		\toprule
		\textbf{Measurement} & \textbf{Bicubic} & \textbf{Bilinear} & \textbf{Nearest Neighbor} \\
		\midrule
		$K$                  & 1.0000           & 1.0000            & 1.0000                    \\
		\bottomrule
	\end{tabular}
\end{table}

\section{Mean of Positional Embedding}
\label{sec:meanpe}

We measured the mean of positional embedding $\E[\bP]$ for ViTs pretrained ImageNet, provided by AugReg \citep{DBLP:journals/tmlr/SteinerKZWUB22}. \tabref{tab:mean} summarizes the results. We conclude that the mean of positional embedding is sufficiently zero-centered. Furthermore, the common initialization for positional embedding is to allow slight variance while using a zero mean. The zero-mean holds for others such as sinusoidal positional embedding. In consideration of this, we assume $\E[\bP] = 0$ in a practical scenario. Here, rescaling positional embedding $\bPte = \UP(\bP)/\sqrt{k}$ yields variance is $\Var[\bP]$ and still exhibits zero-mean.

\begin{table}[h!]
	\caption{Mean of positional embedding}
	\label{tab:mean}
	\centering
	\begin{tabular}{l|ccc}
		\toprule
		\textbf{Model} & \textbf{ViT S/16 $\mathbf{224^2}$} & \textbf{ViT B/16 $\mathbf{224^2}$} & \textbf{ViT L/16 $\mathbf{224^2}$} \\
		\midrule
		$\E[\bP]$      & -0.0014096                         & -0.001129                          & 0.00248424                         \\
		\bottomrule
	\end{tabular}
\end{table}

\section{Notes on Other Data Augmentations}
\label{sec:other}
In the main text, we investigated variance after upsampling, Mixup, Cutmix, and random erasing. However, the standard practice for data augmentation includes other methods. The prime examples are ColorJitter, RandAugment, and AutoAugment. First, for ColorJitter, we claim that the variance shift is in symmetry and its usage is acceptable. If ColorJitter always produces a darker image, we say it is in asymmetry and causes a variance shift. However, ColorJitter can produce both darker and brighter images, which is acceptable. The same goes for contrast and others.

In cases of RandAugment and AutoAugment, there are significant amounts of inner operations, such as AutoContrast, Equalize, Invert, and Posterize, and a subset of them are randomly used in training. Understanding their complicated dynamics would be intractable through equations; in this regard, an empirical approach would be appropriate. Indeed, Reference \citep{DBLP:conf/icml/TouvronCDMSJ21} empirically observed that both RandAugment and AutoAugment improved the performance of ViTs. In consideration of this, we advocate the use of RandAugment and AutoAugment. Nevertheless, because ViTs are sensitive to variance shift, we conjecture that running an automated search to discover the data augmentation can yield a different policy for data augmentations because the current policies were obtained from CNNs. For example, components such as brightness and contrast would be configured to be symmetric because they directly affect the variance of images. Others such as translation would be allowed because they would not affect variance.

\newpage

\section{List of Notations}

\begin{table}[h!]
	\caption{List of notations used in this study.}
	\label{tab:notation}
	\begin{center}
		\begin{tabular}{ll}
			\toprule
			\textbf{Notation}   & \textbf{Meaning}                                                               \\
			\midrule
			$\bI$               & An input image.                                                                \\
			$\bW$               & Convolutional kernel of patch projection.                                      \\
			$*$                 & Strided convolutional operation of patch projection.                           \\
			$\bX$               & Patch embedding.                                                               \\
			$\bP$               & Positional embedding.                                                          \\
			$\LN$               & Layer normalization.                                                           \\
			$\bv$               & A vector.                                                                      \\
			$v_i$               & The $i$th element of vector $\bv$.                                             \\
			$\bv^{\prime}$      & A vector whose elements are permuted.                                          \\
			$\E[\bv]$           & Mean of $\bv$.                                                                 \\
			$\Var[\bv]$         & Variance of $\bv$.                                                             \\
			$(\bv^i ; \bv^j)$   & Concatenated vector from $\bv^i$ and $\bv^j$.                                  \\
			$\Crop$             & Cropping operation.                                                            \\
			$\UP$               & Upsampling operation.                                                          \\
			$N_h, N_w$          & Height, width of input image.                                                  \\
			$N_p$               & Patch size of ViT.                                                             \\
			$k_{1D}, k_{2D}$    & A variance ratio $k = \Var[\UP(\bv)] / \Var[\bv]$ for 1D and 2D upsampling.    \\
			$p$                 & An inference crop percentage.                                                  \\
			$\tbv$              & Mixup sample $\tbv = \lambda \bv^i + (1-\lambda) \bv^j$.                       \\
			$\lambda$           & Mixup ratio.                                                                   \\
			$\bbv$              & Cutmix sample $\bbv = \bM \odot \bv^i + (\bone-\bM) \odot \bv^j$.              \\
			$\bM$               & A binary mask where each element is either zero or one.                        \\
			$\sum\bM$           & Sum of elements in $\bM$, which equals to the length (area) of the alive mask. \\
			$\bv_{:\sum\bM}$    & Subset of vector $\bv$ from the first to ($\sum\bM$)-th element.               \\
			$\bv_{\sum\bM+1:}$  & Subset of vector $\bv$ from ($\sum\bM+1$)-th to the last element.              \\
			$\bone$             & All-ones matrix.                                                               \\
			$\bzero$            & All-zeros matrix.                                                              \\
			$\odot$             & Element-wise multiplication.                                                   \\
			$\hbv$              & Random erasing sample $\hbv = \bM \odot \bv + (\bone-\bM) \odot \be$.          \\
			$\be$               & Erasing vector defined by random erasing mode.                                 \\
			$\pdv{y}{x}$        & Partial derivative of $y$ with respect to $x$.                                 \\
			$\N(\mu, \sigma^2)$ & Normal distribution with mean $\mu$ and variance $\sigma^2$.                   \\
			$\dtbv$             & The extended form of Mixup sample $\dtbv = \lambda_i \bv^i + \lambda_j \bv^j$. \\
			$K$                 & Mean ratio $K = \E[\UP(\bv)] / \E[\bv]$.                                       \\
			\bottomrule
		\end{tabular}
	\end{center}
\end{table}

\end{document}